\documentclass[letterpaper, 10 pt, conference]{ieeeconf}  % Comment this line out if you need a4paper

\IEEEoverridecommandlockouts                              % This command is only needed if 
\title{\LARGE \bf
Towards  Reliable   Colorectal Cancer Polyps Classification via  Vision Based Tactile Sensing and Confidence-Calibrated Neural Networks
}
\author{Siddhartha Kapuria$^{1}$, Tarunraj G. Mohanraj$^{1}$, Nethra Venkatayogi$^{2}$, Ozdemir Can Kara$^{1}$, Yuki Hirata$^{3}$,\\  Patrick Minot$^{4}$, Ariel Kapusta$^{4}$, Naruhiko Ikoma$^{3}$, and Farshid Alambeigi$^{1}$% <-this % stops a space
\thanks{*Research reported in this publication was supported by the University of Texas at Austin and by the MITRE Corporation through the MITRE Innovation Program. Approved for Public Release; Distribution Unlimited. Public Release Case Number 23-0438.}% <-this % stops a space
\thanks{$^{1}$Siddhartha Kapuria,  Tarunraj G. Mohanraj, Ozdemir Can Kara and Farshid Alambeigi are with Walker Department of Mechanical Engineering, University of Texas at Austin, Austin, TX, USA. email: {\tt\small \{skapuria, tarunrajgm, ozdemirckara\}@utexas.edu, and farshid.alambeigi@austin.utexas.edu}}%
\thanks{$^{2}$Nethra Venkatayogi is with the Department of Biomedical Engineering, University of Texas at Austin, Austin, TX, USA. email: {\tt\small venkatayoginethra@utexas.edu}
}
\thanks{$^{3}$Naruhiko Ikoma and Yuki Hirata are with the Department of Surgical Oncology, Division of Surgery, The University of Texas MD Anderson Cancer Center, Houston, TX, USA, 77030. email: {\tt\small \{nikoma,yhirata\}@mdanderson.org}}
\thanks{$^{4}$Patrick Minot and Ariel Kapusta are with the MITRE Corporation.}%
\thanks{The dataset used in this study has been made publicly available in a Github repository named Trustworthy AI for CRC Polyp Detection (\url{https://github.austin.utexas.edu/mech-arts-lab/Trustworthy-AI-for-CRC-Polyps-Detection})}
}

\let\labelindent\relax
\usepackage{enumitem}
\usepackage{graphicx}
\graphicspath{ {./Figures/} }
\usepackage{subcaption}
\usepackage{multirow}
\usepackage{hyperref}
\hypersetup{
    colorlinks=true,
    linkcolor=blue,
    filecolor=magenta,      
    urlcolor=cyan,
    }

\usepackage{fancyhdr}
\fancypagestyle{firstpage}{%
  \lhead{This paper has been accepted for publication at the 2023 International Symposium on Medical Robotics Conference.}
}

\usepackage{color}

\begin{document}

\maketitle
\thispagestyle{firstpage}

%%%%%%%%%%%%%%%%%%%%%%%%%%%%%%%%%%%%%%%%%%%%%%%%%%%%%%%%%%%%%%%%%%%%%%%%%%%%%%%%
\begin{abstract}
In this study, toward addressing the  over-confident outputs of existing artificial intelligence-based colorectal cancer (CRC) polyp  classification techniques, we propose a confidence-calibrated residual neural network. Utilizing a novel vision-based tactile sensing (VS-TS) system and unique CRC polyp phantoms, we demonstrate that  traditional metrics such as accuracy and precision are not sufficient to encapsulate
model performance  for handling a sensitive CRC polyp diagnosis. To this end, we 
develop a residual neural network classifier and address its
over-confident outputs for CRC polyps classification via the
post-processing method of temperature scaling.
To evaluate the proposed method,  we introduce noise and blur to the obtained textural images of the VSTS and  test the model's reliability for non-ideal inputs  through reliability diagrams and other statistical metrics.  %\ari{Assured is a overused and underdefined term. I'm struggling to create a definition. Something like "Assurance = A process spanning the entire product development lifecycle that can guarantee, to a defined level, system safety and effectiveness". Improving confidence estimations is towards "assurance", but cavalier use of the term muddies its meaning.} 

\end{abstract}

%%%%%%%%%%%%%%%%%%%%%%%%%%%%%%%%%%%%%%%%%%%%%%%%%%%%%%%%%%%%%%%%%%%%%%%%%%%%%%%%
\section{INTRODUCTION}

Colorectal cancer (CRC) is the third most diagnosed cancer in the United States \cite{Sung2021GlobalCS}. Early detection of pre-cancerous polyp lesions  can potentially increase the survival rate of patients to almost 90\% \cite{Jemal2002CancerS2}. It has been shown that the morphological characteristics of CRC polyps observed during colonoscopy screening can be used as an indicator of the neoplasticity of a polyp (i.e. its cancerous potential)  \cite{Li2009MacroscopicBT}, \cite{Axon2005UpdateOT}. However, the task of early CRC polyp detection and classification using colonoscopy images is highly complex and clinician-dependent \cite{Lou2014ARS}, increasing the risk of early detection miss rate (EDMR) and mortality.

To address the critical EDMR  issue, computer-aided diagnostics using artificial intelligence (AI) has increasingly been employed for improving the detection and characterization of cancer polyps. 
% A set of guidelines to classify polyps under different “\textit{pit-pattern}'' structures on their surface was first introduced by Kudo et al. \cite{Kudo1994ColorectalTA}.
\begin{figure}[t]
		\centering
		% \hspace*{-8mm}	
		\includegraphics[scale=0.28]{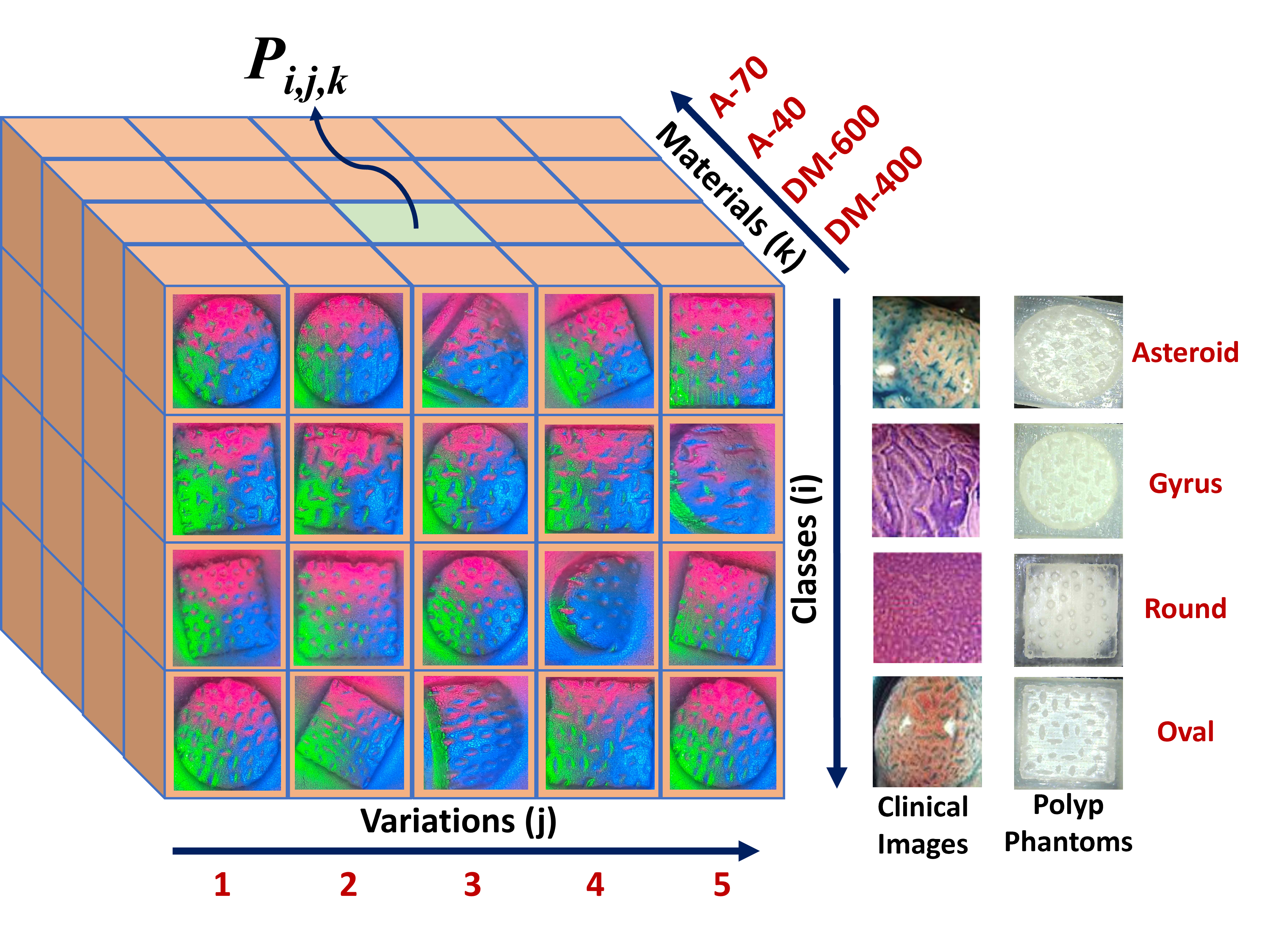}
		\vspace*{-2mm}
\caption{A conceptual three-dimensional illustration of the image dataset collected by  HySenSe on 5 out of the 10 variations of 3D printed polyp phantoms. These polyp phantoms are classified based on Kudo pit-patterns, such as Asteroid, Gyrus, Round, and Oval~\cite{Kudo1994ColorectalTA}. Moreover, each of the polyp phantoms is printed with 4 different materials (i.e., DM-400, DM-600, A-40, and A-70.}
\label{fig:cube}
    \vspace*{-8mm}
\end{figure}
% Since the task of early CRC polyp detection and classification is highly complex and observer-dependent \cite{Lou2014ARS}, computer-aided diagnostic (CAD) using artificial intelligence (AI) have increasingly been employed for improving the detection and characterization of cancer polyps. 
Examples of the utilized AI algorithms include support vector machines (SVM), k-nearest neighbors (k-NN), ensemble methods, random forests, and convolutional neural networks (CNN) \cite{Kara2023abme, Wang2021ArtificialID,Shin2017ComparisonOH,Viscaino2021ArtificialIF}. Due to the difficulties associated with obtaining medical data and patient records to generate datasets, recently, transfer learning using neural networks pre-trained on large general-purpose datasets such as ImageNet \cite{Deng2009ImageNetAL} has also become  a widely popular technique to aid in medical computer aided diagnostics, and in particular, the detection and classification of CRC polyps \cite{Ribeiro2016ExploringDL}, \cite{Zhang2017AutomaticDA}. 
To evaluate the performance of the utilized AI algorithms, statistical metrics such as accuracy, precision, sensitivity, and recall are typically used in the literature. For example, Zhang et al. \cite{Zhang2017AutomaticDA} used precision, accuracy, and recall rate to evaluate the performance of the implemented AI algorithms, while Ribeiro et al. \cite{Ribeiro2016ExploringDL} only used accuracy as an evaluation metric. 

A review of the literature demonstrates that using the aforementioned statistical metrics, researchers  mainly have focused on the ``\textit{correctness}" of the predictions and not  the ``\textit{reliability}" and ``\textit{confidence}" of the implemented AI algorithms. In other words, these studies solely have focused on comparing the correctness of the predicted labels with the ground truth labels.
% \ari{I might suggest saying that in sensitive applications it is critical to reduce incorrect diagnosis and that a more accurate confidence estimate can better inform clinicians basing decisions on the diagnosis. This description essentially is the reason for the criticality of the confidence estimation. Also note that the other methods have something that they might call confidence, but it is poorly related to correctness or reliability (which is what the next paragraph talks about).}
Nevertheless, in sensitive AI applications such as cancer diagnosis,  it is also critical to reduce incorrect diagnoses by reporting the likelihood of correctly predicting the labels, through attaching a ``confidence" metric to each prediction.
% Nevertheless, in sensitive AI applications such as cancer diagnosis,  it is also critical to attach a ``confidence" metric to each prediction and determine the  likelihood of correctly  predicting the labels.
% Of note, this important metric thus describes the reliability of the predictions by including the probability of the prediction being correct.
Of note, accurately providing a confidence level significantly improves the interpretability and appropriate level of trust of the model's output. For instance, for the case of CRC polyps' detection and classification, 
% when confidence level of a prediction is low, clinician's decision  and diagnosis would be critical to avoid a potential EDMR and diagnostic errors. 
a more accurate confidence estimate can better inform clinicians basing decisions on the AI diagnosis.

In case of deep neural networks, it is often erroneously assumed that the output of the final classification layer (i.e., softmax) is a realistic measure of confidence \cite{Gal2016UncertaintyID}. However, as shown by Guo et al. \cite{Guo2017OnCO} taking the example of a ResNet with 110 layers \cite{He2016DeepRL}, deep neural networks often produce a higher softmax output than the ground truth demonstrating over-confident results. Such a network with a difference in ground truth probabilities and the predicted softmax outputs is called a ``\textit{miscalibrated}" network \cite{Guo2017OnCO}. To address this miscalibration issue and use softmax outputs of neural networks as realistic confidence estimates, different techniques have been explored in the literature. For example,  Guo et al. \cite{Guo2017OnCO} provided insights on simple post-processing calibration methods to obtain accurate confidence estimates. Moreover, modifying the loss function  using the difference between confidence and accuracy (DCA) \cite{Liang2020ImprovedTC} and Dynamically Weighted Balanced (DWB) \cite{Fernando2022DynamicallyWB} have also been explored by researchers. 
Similar efforts have been made in the medical imaging community to incorporate confidence calibration in neural network models. For instance, Carneiro et al. \cite{Carneiro2020DeepLU} explored the role of confidence calibration for  polyps classification based on  colonoscopy images and used the temperature scaling technique for network calibration. Building on this, Kusters et al. \cite{Kusters2022ColorectalPC} employed trainable methods based on DCA and DWB for confidence calibration. 

\begin{figure}[t!]
		\centering
		% \hspace*{-8mm}	
  		\vspace*{2mm}
		\includegraphics[scale=0.3]{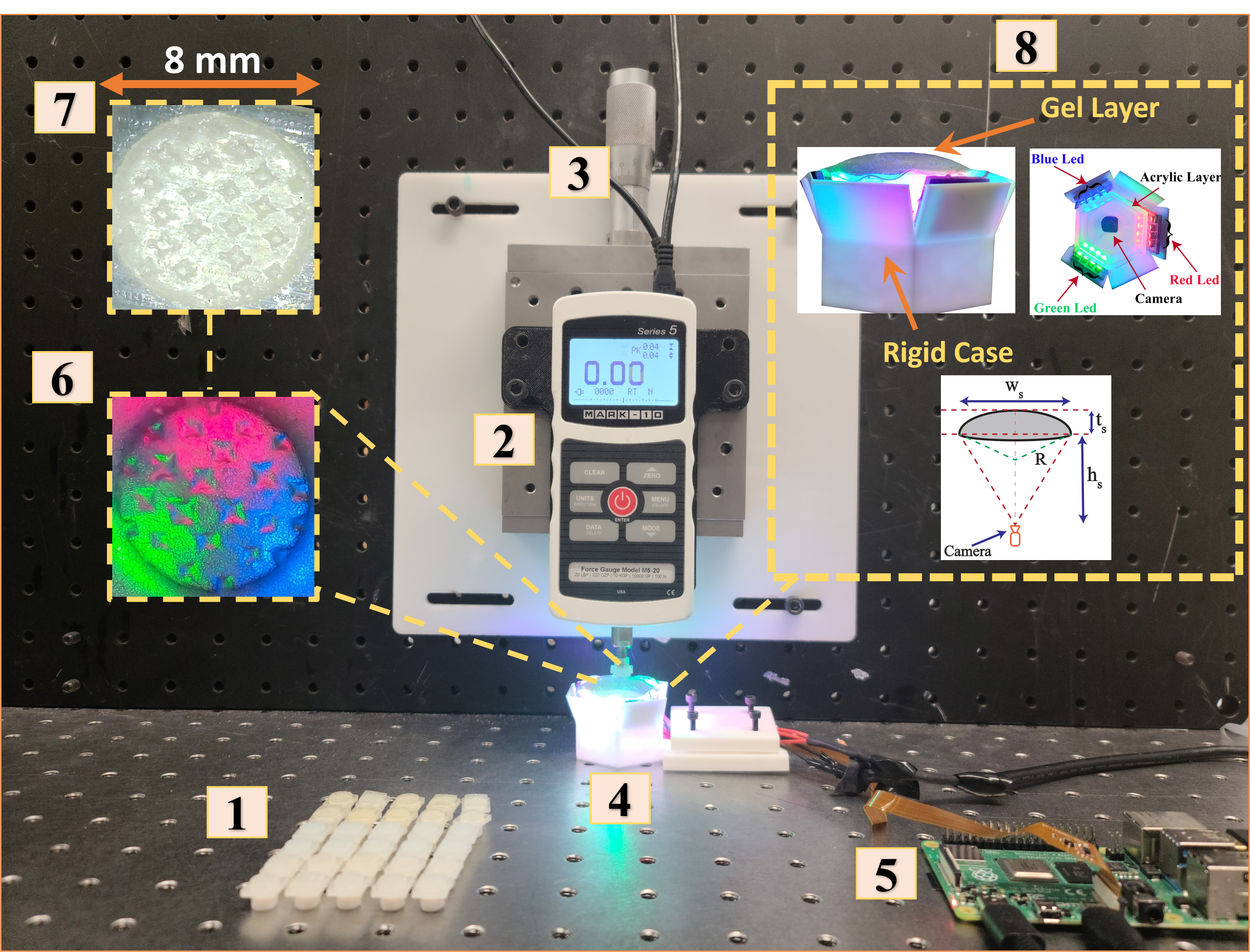}
		\vspace*{-2mm}
\caption{Experimental Setup: \textcircled{1}: CRC polyp phantoms, \textcircled{2}: Mark-10 Series 5 Digital Force Gauge, \textcircled{3}: M-UMR12.40 Precision Linear Stage, \textcircled{4}: HySenSe sensor, \textcircled{5}: Raspberry Pi 4 Model B, \textcircled{6}: HySenSe image output, \textcircled{7}: Dimensions of polyp phantom, \textcircled{8}: HySenSe side view, top view, and dimensions, $h_s$ = 24 mm is the height of the 3D printed rigid frame, $t_s$ = 4.5 mm is the thickness of the gel layer $w_s$ = 33.8 mm is the width of the gel layer, and R = 35 mm is the radius of the dome-shaped gel layer.}
\label{fig:setup}
    \vspace*{-8mm}
\end{figure}

In our recent work \cite{Venkatayogi2022ClassificationOC}, solely utilizing the typical evaluation metrics (i.e., precision, accuracy, and recall), we demonstrated the high potentials of utilizing a dilated Convolutional Neural Network (CNN) to precisely and sensitively  (i.e., an average accuracy of 93\%) classify CRC polyps under the Kudo classification system \cite{Kudo1994ColorectalTA}. Unlike common images provided during colonoscopy screening, this framework utilizes unique 3D textural images (shown in Fig. \ref{fig:cube}) captured by the HySenSe sensor \cite{Kara2022HySenSeAH}, which is a novel hyper sensitive and high fidelity vision-based surface tactile sensor (VS-TS). 
In this paper, towards developing a reliable and interpretable CRC polyp classification model and in order to address our over-confident results  in \cite{Venkatayogi2022ClassificationOC}, we further develop our best-performing ML classifier and address its confidence calibration for CRC polyps classification via the post-processing method of temperature scaling. We also focus on improving the model's generalization ability for non-ideal inputs (i.e.,  noisy and blurry textural images) and calculating the likelihood of the CRC polyps' prediction to the true class.
% to  improve its output's interpretability and reliability.  

% Using the post-processing method of temperature scaling, we now focus on improving the model's generalization ability for non-ideal inputs (i.e., a noisy and blurry textural image) and attaching a real-world estimate of the likelihood of the prediction to the true class.
% Of note, this analysis provides another layer of safety to reduce EDMR, such that the final decision on a low-confidence prediction could be more closely evaluated by skilled clinicians, while a high-confidence prediction would incorporate a considerable level of reliability to the diagnosis. 

\begin{figure}[t!]
\vspace*{1.5mm}\centering
  \begin{subfigure}[b]{0.48\textwidth}
    \includegraphics[width=1\textwidth]{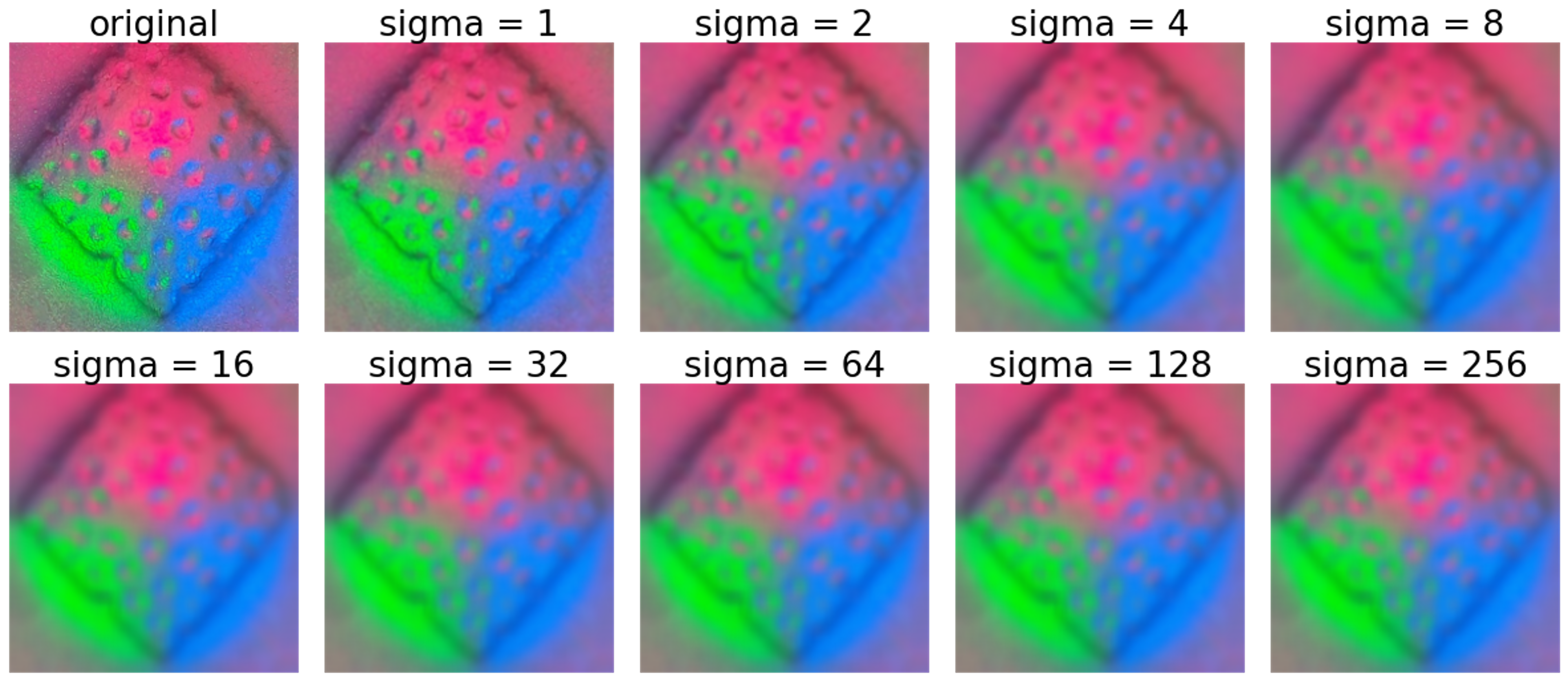}
    \caption{Blur}
    \label{fig:blur_levels}
  \end{subfigure}
  \hfill
  \begin{subfigure}[b]{0.48\textwidth}
    \includegraphics[width=1\textwidth]{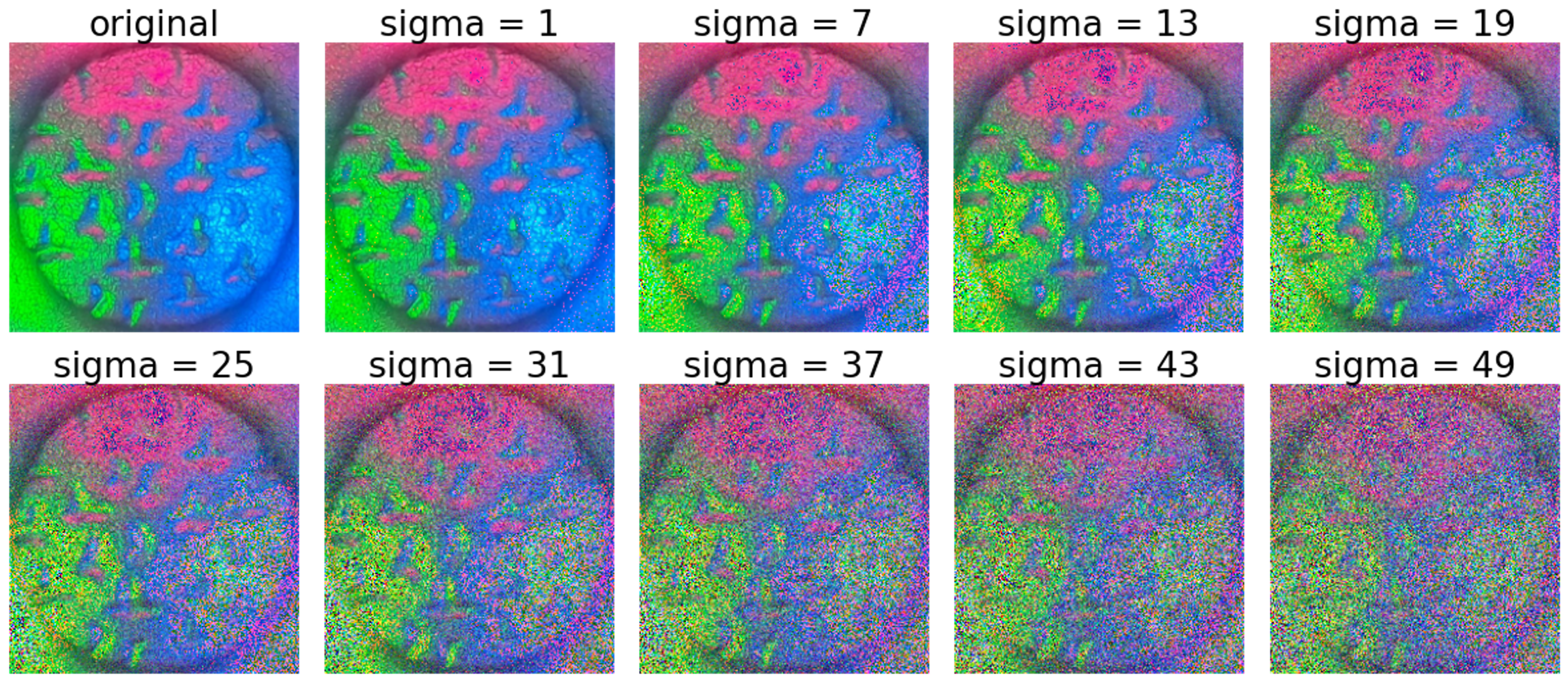}
    \caption{Noise}
    \label{fig:noise_levels}
  \end{subfigure}

\caption{Exemplary visuals for noise and blur levels that are utilized in the datasets.} 
\label{fig:levels}
\vspace*{-8mm}
\end{figure}
\vspace*{-3mm}\section{MATERIALS AND METHODS}

\subsection{Vision Based Tactile Sensor (VS-TS)}

In this study, we utilized  a novel VS-TS called HySenSe developed in~\cite{Kara2022HySenSeAH} to collect high-fidelity textural images of CRC polyp phantoms for the training and evaluation of our confidence-calibrated AI model. As shown in Fig. \ref{fig:setup}, this sensor consists of: (I) a deformable silicone membrane that directly interacts with polyp phantoms, (II) an optical module (Arducam 1/4 inch 5 MP camera), which captures the tiny deformations of the gel layer when there exist an interaction with a polyp phantom, (III) a transparent acrylic plate providing support to the gel layer, (IV) an array of Red, Green and Blue LEDs to provide internal illumination for the depth perception, and (V) a rigid frame supporting the entire structure. Working principle of the VS-TS is very simple yet highly intuitive in which the deformation caused by the interaction of the deformable membrane with the CRC polyps surface can visually be captured by a camera embedded in the frame. More details about the fabrication and functionality of this sensor can be found in~\cite{Kara2022HySenSeAH}. 

\subsection{Polyp Phantoms and Experimental Procedure}

% We employed the same dataset as in our previous study~\cite{Venkatayogi2022PitPatternCO}. 
Fig.~\ref{fig:cube} illustrates a 3D tensor conceptually illustrating the fabricated CRC polyp phantoms designed and additively manufactured based on the realistic CRC polyps described in~\cite{Kudo1994ColorectalTA}. As shown in this figure, 
by varying the indices ($i$, $j$, $k$) along each side of the tensor, a unique polyp $P_{i, j, k}$ can be characterized showing one of four Kudo pit-pattern classifications $i$ (referred to as A (Asteroid), G (Gyrus), O (Oval/Tubular), and R (Round) throughout this paper)~\cite{Kudo1994ColorectalTA}, one of ten geometric variations $j$, and one of four materials with different hardness $k$ (representing different stages of cancer ~\cite{zanotelli2018mechanical}).
% Of note, classes A and R are representing non-neoplastic (i.e., non-cancerous), while classes O and G are among the neoplastic variants \cite{rex2019optimal}.
Across the four classes, the feature dimensions range from 300 to 900 microns, with an average spacing of 600 microns between pit patterns. Following the design of the polyp phantoms CAD model in SolidWorks (SolidWorks, Dassault Systemes), each of the 160 unique  phantoms that constitute the dataset was printed with the J750 Digital Anatomy Printer (Stratasys, Ltd) with different material combinations shown in Fig. \ref{fig:cube}. More details about the fabrication of polyp phantoms can be found in \cite{Venkatayogi2022ClassificationOC}.

% set of 160 unique CRC polyp phantoms across four different pit-pattern classes of the Kudo classification system [CITE] formed the basis of the dataset. Each polyp is characterized by one of four Kudo classes (referred to as Asteroid (A), Gyrus (G), Oval (O), and Round (R) throughout this paper, respectively), one of ten geometric textural variations within a class, and one of four materials, which represent varying stiffness qualities. Following the design of polyp phantoms in SolidWorks (SolidWorks, Dassault Systemes), each of the 160 polyps was printed using the J750 Digital Anatomy Printer (Stratasys, Ltd) with different material combinations. 
\vspace*{-1.7mm}\subsection{Experimental Setup and Data Collection Procedure}
Utilizing the Hysense sensor, CRC polyp phantoms, and the experimental setup shown in Fig.~\ref{fig:setup},  a set of experiments were conducted  under two contact angles of 0° and 45° between the polyp face and HySenSe. Of note, 0° mimics a complete interaction between the HySenSe deformable layer and the polyp, in which the whole texture of the polyp can be captured by HySenSe, whereas the 45° simulates a case in which limited portion of the polyp’s texture can be captured by the sensor. Each of the 160 unique polyps had an interaction with the HySenSe until a 2 N force was exerted in the 0° orientation. Additionally, five out of the ten geometric variations $j$ chosen randomly from each polyp class $i$, across each of the four materials $k$, were used for the experiments in the 45° orientation, for a total of 80 angled experiments resulting a total of 229 samples that constitute the  dataset.

\vspace*{-1.7mm}\subsection{Datasets and Pre-Processing}
Between 229 samples, the class counts for each polyp A, G, O, and R were 57, 57, 55, and 60, respectively. From the 229 polyp visuals, training via stratified K-fold cross-validation was performed with 5 folds on 80\% (182 samples) of the dataset, while 20\% (47 samples) was reserved for model evaluation purposes. The obtained HySenSe visuals were manually cropped to only include the polyp area of interest and downsized from the native 1080 × 1280 pixels to 224 × 224 to improve model performance. Three different datasets were constructed  using the same training data split: (I) with neither Gaussian Blur nor Gaussian Noise transformations on the base samples (examples shown in Fig. \ref{fig:cube}), (II) with Gaussian Blur (Fig. \ref{fig:blur_levels}), and (III) with Gaussian Noise (Fig. \ref{fig:noise_levels}). %, and (IV) with both blur and noise transforms. 
Of note, the Gaussian transforms used in datasets (I)-(III) occur at a probability of 0.5 for each sample. We also used a value of $\sigma$ ranging from 1 to 256 for blur, and a $\sigma$ of 1 to 50 for adding the noise values. Notably, higher values of $\sigma$ denote more significant blur and noise, as illustrated in Fig.~\ref{fig:levels}. To improve generalization, we chose the maximum blur and noise limits to be well beyond the worst case that the model may encounter in a clinical setting. Additionally, to further improve the model's robustness, all four sets included standard geometric augmentations, such as random cropping, horizontal and vertical flips, and random rotations between -45° and 45°, each with an independent occurrence probability of 0.5. 

The original 47 samples (i.e. 20\% of the dataset) that were reserved for model evaluation were used to construct an expanded test set consisting of a total of $47 \times 4= 188$ samples. This was achieved by combining four independent, visually distinct groups of the same 47 samples: Group A consisting of "clean" images without any Gaussian transformations applied to the samples, Group B with each of the 47 samples incorporating varying levels of Gaussian Blur, Group C with each of the samples incorporating varying levels of Gaussian Noise, and Group D with all the images experiencing a combination of both Gaussian Blur and Gaussian Noise. The 188 images resulting from combining the samples in Group A-D were used to evaluate the calibration performance of the model trained on Datasets I-III. To simulate a more reasonable level of blur and noise in the test set that the model may encounter in a real-world setting, we limited the maximum values of $\sigma$ to 32 and 30 for blur and noise, respectively.

% The base test set was developed into 188 samples by combining four visually distinct groups of the same 47 images: Group A consisting of clean images without Gaussian transforms, Group B with all images incorporating Gaussian Blur, Group C with all images incorporating Gaussian Noise, and Group D with all images experiencing a combination of blur and noise. For the test set, to simulate a more reasonable level of blur and noise that the model would be expected to encounter in a real-world scenario, we limited the maximum values of $\sigma$ to 32 and 30 for blur and noise, respectively. The models trained on Datasets I-III were all evaluated on this test set. 

% To test the model's capabilities to generalize at higher blur and noise levels, we also plot the accuracy and confidence of the calibrated and uncalibrated model trained on each dataset from I to III versus increasing blur and noise. Here again, we use a value of $\sigma$ ranging from 1 to 256 for blur and from 1 to 50 for noise, with a logarithmic step size for blur and a linear step for noise. 

% since that is a more reasonable expectation of the kind of images the model might encounter. 

%%talk about differences in blur/noise ranges for test v. training 

\vspace*{-2mm}\subsection{Model Architecture}
The residual network (ResNet) architecture is the current standard ML model for polyp classification tasks \cite{patel2020comparative}, \cite{wang2021multiclassification} due to its ability to curtail exploding gradients \cite{he2016deep}, \cite{huang2020deep}. Additionally, ResNets use skip connections to lessen the degradation problem, where model performance is negatively impacted by increasing its complexity \cite{he2016deep}. Notably, standard ResNet convolutional layers do not utilize dilated kernels. Dilations maintain the spatial resolution of feature maps encountered during convolutions while also enhancing the network’s receptive field to observe more details. In \cite{Venkatayogi2022PitPatternCO}, we demonstrated the effectiveness of using a dilated CNN---inspired by the ResNet architecture---to capture and classify the intricate textural features seen in our dataset, while outperforming state-of-the-art networks across a wide range of clinically relevant metrics. In this work, we employ the same model to examine its response to calibration and address the over-confident results reported in \cite{Venkatayogi2022PitPatternCO}. 

% As described in [CITE ICRA], we use a ResNet architecture that incorporates dilated convolutions to enhance performance without increased model complexity and simultaneously deter overfittingon our small dataset. These dilations will increase the convolution layers' receptive field to capture intricate contextual details, which are critical to reliable polyp classification.

\begin{table*}[t]
\vspace*{2mm}
  % \centering
  \begin{center}
  \caption{\label{tab:results} Calibration Results}
  \begin{tabular}{|p{0.8cm}||p{1.4cm}|p{1.4cm}||p{1.4cm}|p{1.4cm}||p{1.4cm}|p{1.4cm}||p{1.4cm}||p{1.4cm}||p{1cm}| }
    \hline
    \multirow{2}{1em}{Dataset} & \multicolumn{2}{|c||}{ECE} & \multicolumn{2}{|c||}{MCE} & \multicolumn{2}{|c||}{ACE} & \multicolumn{2}{|c||}{Average Confidence} & \multirow{2}{1em}{Accuracy}\\
    \cline{2-9}
    & Uncalibrated & Calibrated & Uncalibrated & Calibrated & Uncalibrated & Calibrated & Uncalibrated & Calibrated\\
    \hline
    I & 0.187 & 0.0901 & 0.346 & 0.291 & 0.211 & 0.139 & 79\% & 61\% & 60\% \\
    \hline
    II & 0.166 & 0.093 & 0.385 & 0.119 & 0.184 & 0.0845 & 82\% & 64\% & 62\% \\
    \hline
    III & 0.124 & 0.0663 & 0.271 & 0.258 & 0.169 & 0.090 & 78\% & 64\% & 68\% \\
    \hline
%    IV & 0.0624 & 0.0513 & 0.191 & 0.201 & 0.0957 & 0.0825 & 79\% & 80\% & 80\% \\
%    \hline
  \end{tabular}
  \end{center}
\vspace*{-6mm}
\end{table*}
\begin{figure*}[t]
\centering
\begin{minipage}{.33\textwidth}
   \begin{subfigure}[b]{1\textwidth}
    \includegraphics[width=1\textwidth]{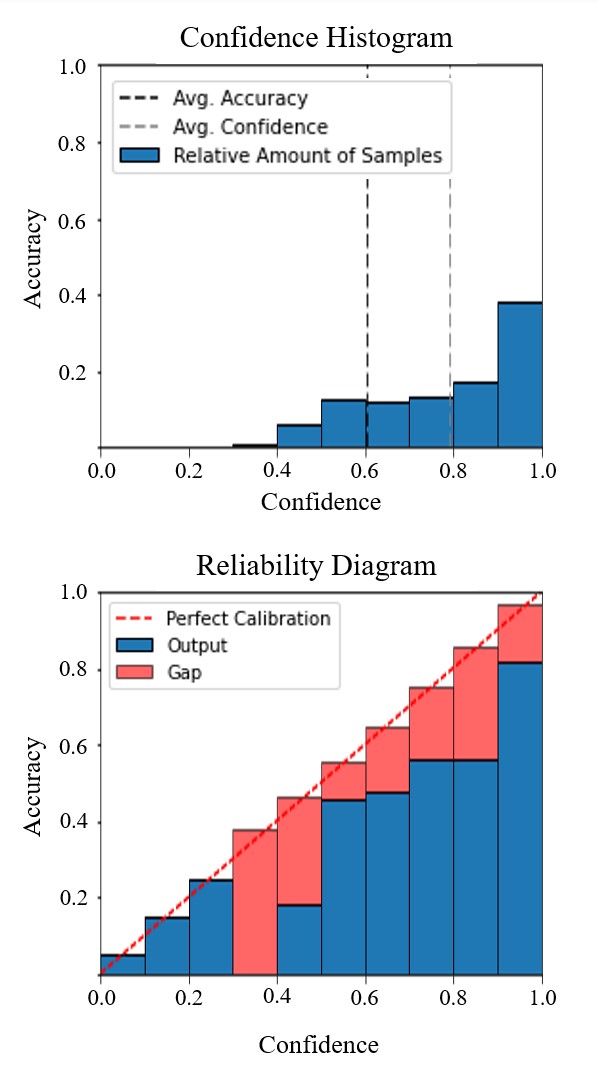}
    \caption{Uncalibrated}
    \label{fig:dataset1_uncalib}
  \end{subfigure}
    \hspace{1em}%
     \begin{subfigure}[b]{1\textwidth}
    \includegraphics[width=1\textwidth]{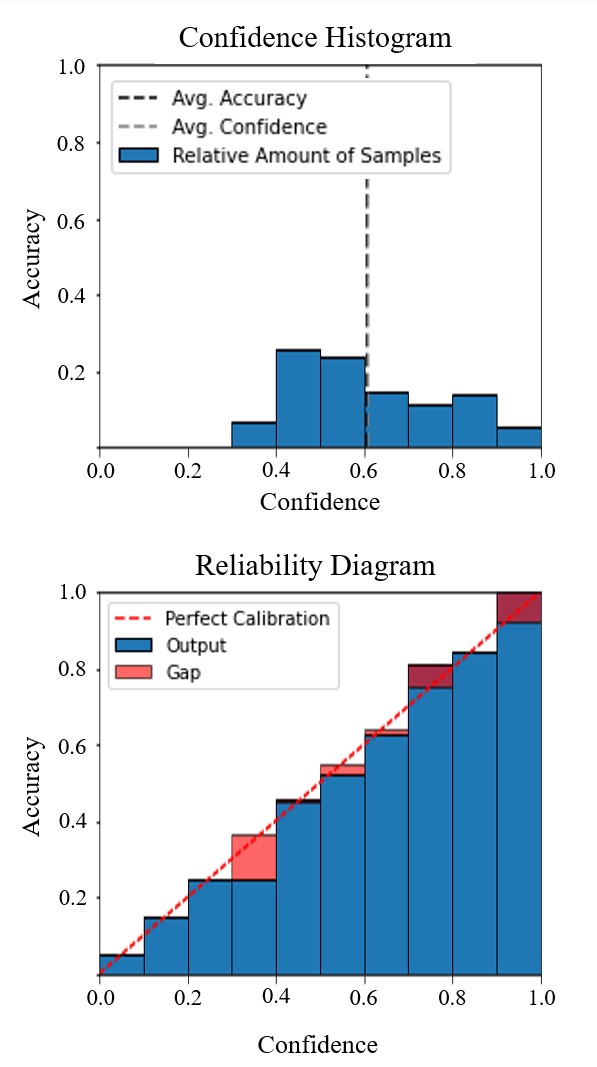}
    \caption{Calibrated}
    \label{fig:dataset1_calib}
  \end{subfigure}
    \caption{\centering Model trained on Dataset I} 
    \label{fig:plots1}
\end{minipage}%
\hfill
\begin{minipage}{.33\textwidth}
   \begin{subfigure}[b]{1\textwidth}
    \includegraphics[width=1\textwidth]{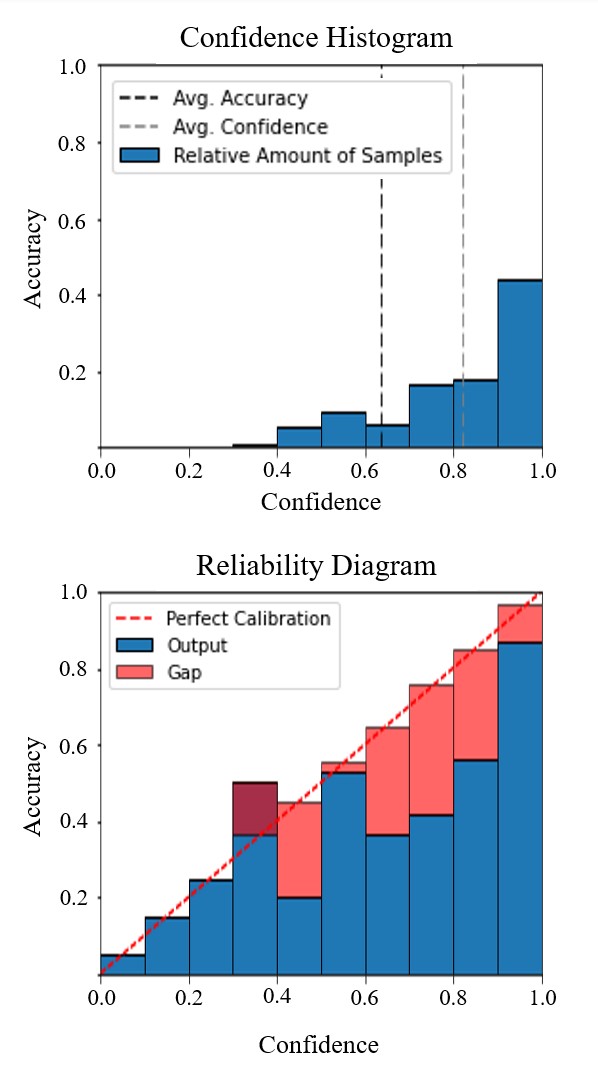}
    \caption{Uncalibrated}
    \label{fig:dataset2_uncalib}
  \end{subfigure}
    \hspace{1em}%
     \begin{subfigure}[b]{1\textwidth}
    \includegraphics[width=1\textwidth]{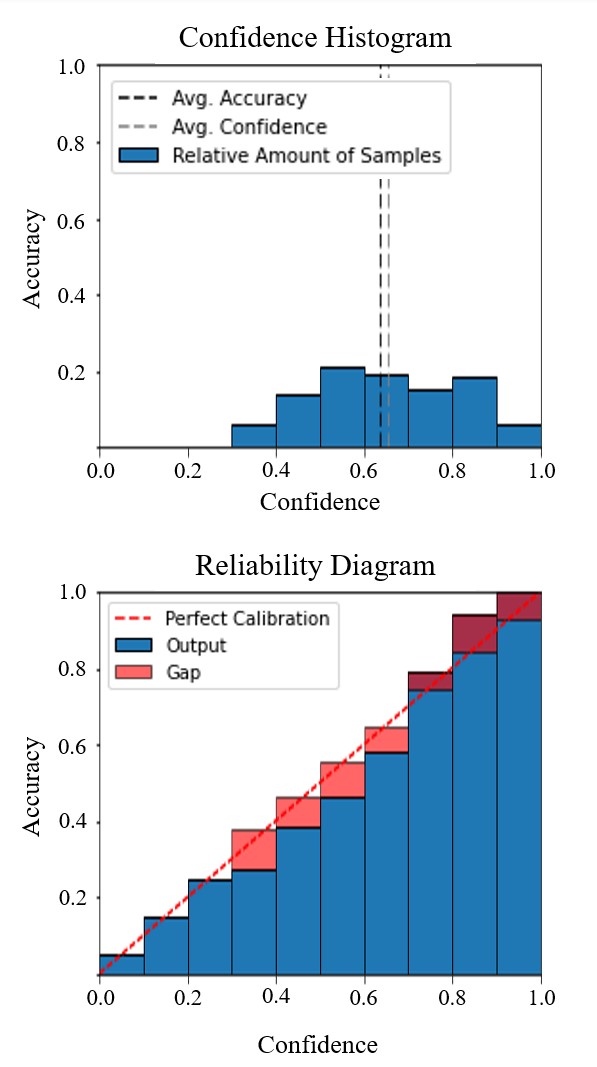}
    \caption{Calibrated}
    \label{fig:dataset2_calib}
  \end{subfigure}
    \caption{\centering Model trained on Dataset II} 
    \label{fig:plots2}
\end{minipage}%
\hfill
\begin{minipage}{.33\textwidth}
   \begin{subfigure}[b]{1\textwidth}
    \includegraphics[width=1\textwidth]{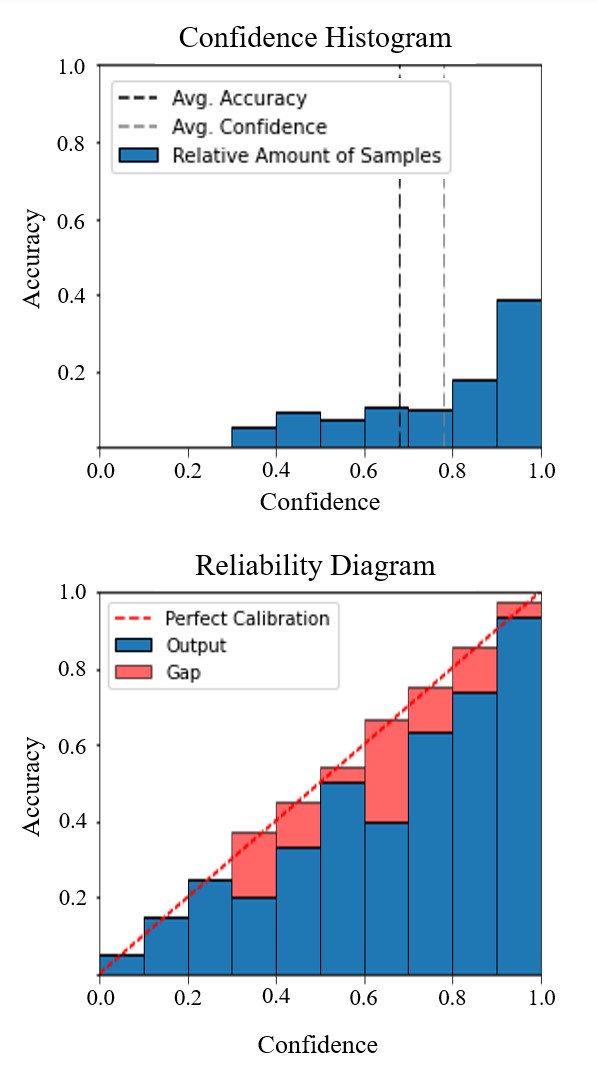}
    \caption{Uncalibrated}
    \label{fig:dataset3_uncalib}
  \end{subfigure}
    \hspace{1em}%
     \begin{subfigure}[b]{1\textwidth}
    \includegraphics[width=1\textwidth]{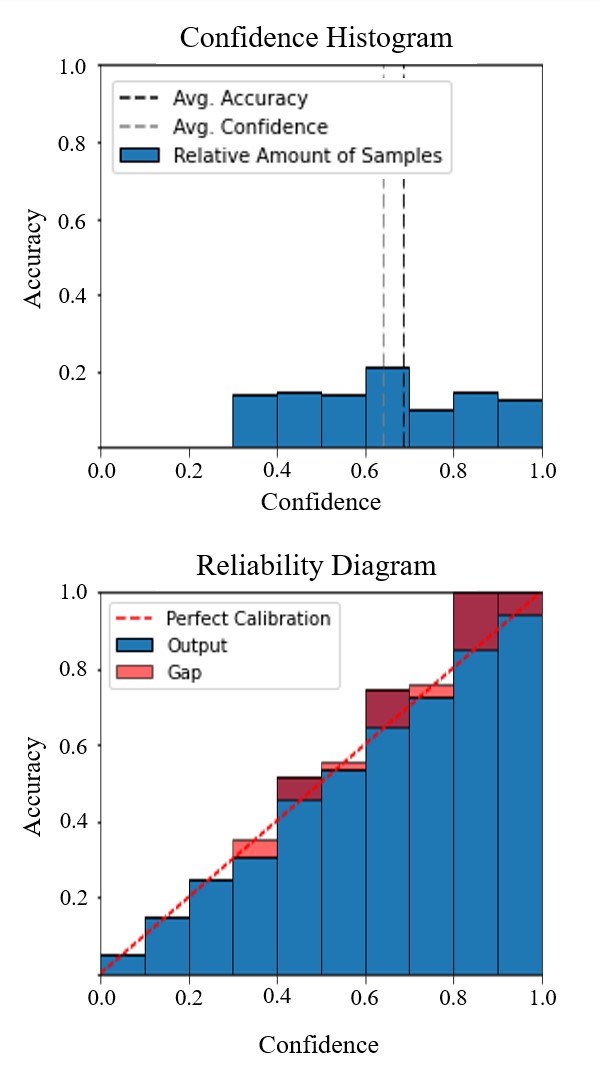}
    \caption{Calibrated}
    \label{fig:dataset3_calib}
  \end{subfigure}
    \caption{\centering Model trained on Dataset III} 
    \label{fig:plots3}
\end{minipage}%
\end{figure*}

%Wider Images

\vspace*{-2mm}\subsection{Model Calibration}

Confidence calibration is the problem of matching the output confidence level with the actual likelihood of the model. It is an important step towards improving model interpretability as most of the deep neural networks are typically over-confident in their predictions \cite{Guo2017OnCO}. Of note, A model is said to be perfectly calibrated when the confidence level of a prediction represents the true probability of the prediction being correct \cite{Guo2017OnCO}. Mathematically speaking, if input $X$ is considered with class labels $Y$, the predicted class is $\hat{Y}$ and $\hat{P}$ is its associated confidence, then for perfect calibration, the probability $\mathbf{P} is$:
$$ \mathbf{P}(\hat{Y} = Y | \hat{P} = p) = p,  \;\; \forall \;p  \in [0, 1]$$
 where the probability is over the joint distribution.

%\begin{figure*}[t]
%\centering
%  \begin{subfigure}[b]{0.48\textwidth}
%    \includegraphics[width=1\textwidth]{Figures/mixed_uncalib.png}
%    \caption{Uncalibrated}
%    \label{fig:dataset4_uncalib}
%  \end{subfigure}
%  \hfill
%  \begin{subfigure}[b]{0.48\textwidth}
%    \includegraphics[width=1\textwidth]{Figures/mixed_calib.png}
%    \caption{Calibrated}
%    \label{fig:dataset4_calib}
%  \end{subfigure}
%
%\caption{\centering Training on Dataset IV} 
%\label{fig:plots4}
%\end{figure*}

\subsubsection{Temperature Scaling}
Temperature scaling is the simplest extension of Platt scaling \cite{Platt1999ProbabilisticOF} and uses a single parameter $T>0$ for all cases. 
Guo et al. \cite{Guo2017OnCO} have shown that temperature scaling is an effective method for confidence calibration. Although other trainable calibration methods (e.g., DCA and DWB)  also exist, we chose temperature scaling due to  its simplicity and independence from model training. Given a logit vector $z_i$ which is the input to the SoftMax function $\sigma_{SM}$, the new confidence prediction is:  

$$\hat{q_i} = \max_k \sigma_{SM} \left(\frac{z_i}{T}\right)^{(k)} $$

where, parameter $T$ is the temperature, and is learned over the holdout validation set by minimizing the negative log-likelihood. Of note, this approach works by “softening” out the output SoftMax function.

\subsubsection{Reliability Diagrams}

Reliability diagrams are an intuitive way of visually representing model calibration \cite{Guo2017OnCO}. By grouping predictions into bins based on their confidence levels and calculating the average accuracy in each bin, we can plot the expected sample accuracy as a function of confidence. The diagram of a perfectly calibrated model plots the identity function, and gaps in calibration can be seen as the deviation from the identity function. 

Taking M equal spaced confidence bins with $B_m$ to be the set of indices in the $m^{th}$ confidence interval, and $n$ the number of samples, the accuracy of $B_m$ can be calculated as \cite{Guo2017OnCO} \cite{Kusters2022ColorectalPC}: 

$$ acc(B_m) = \frac{1}{|B_m|} \sum_{i\in B_m} 1(\hat{y_i}=y_i) $$

The average confidence within $B_m$, taking $\hat{p_i}$ to be the confidence of sample $i$ is\cite{Guo2017OnCO}, is: 

$$ conf(B_m) = \frac{1}{|B_m|} \sum_{i\in B_m} \hat{p_i} $$

\subsubsection{Metrics}

 In addition to accuracy (A), sensitivity (S), and precision (P), we use the following scalar summary statistics for calibration. Similar to reliability diagram construction, the predictions are divided into M-equal confidence bins. Taking $B_m$ to be the set of indices in the $m^{th}$ confidence interval, and $n$ the number of samples, we have:

\begin{enumerate}[label=\alph*)]
\item
MCE: Maximum Calibration Error \cite{Guo2017OnCO}:
$$ MCE = \max_{m \in \{1, 2, ... M\}}|acc(B_m) - conf(B_m)|$$

\item
ECE: Expected Calibration Error \cite{Guo2017OnCO}:

$$ ECE = \sum_{m=1}^{M} \frac{|B_m|}{n}|acc(B_m) - conf(B_m)|$$

\item
ACE: Average Calibration Error \cite{Guo2017OnCO}:
$$ ACE = \frac{1}{M^+} \sum_{m=1}^{M}|acc(B_m) - conf(B_m)|$$

where $M^+$ is the number of non empty bins.
\end{enumerate}

\section{RESULTS AND DISCUSSION}

Accuracy, Average Confidence, MCE, ECE, and ACE were recorded for the network trained on Datasets (I)-(III) and evaluated on a dataset that incorporates clean, blurry, and noisy images, as well as a combination of noise and blur. Results have been summarized in Table \ref{tab:results} and shown in Figs. \ref{fig:plots1}-\ref{fig:plots3}. As can be observed from these results,  the calibrated models produce lower MCE, ECE, and ACE, although the gap varies between the datasets. It is of note that when trained on Dataset I (i.e., Fig. \ref{fig:plots1}), which contains only clear images, the model accuracy over the test set (which contains blurry and noisy images) is only 60\%, yet the average reported confidence is 80\%. This considerable discrepancy between the model's true performance (i.e. accuracy) and its reported performance (i.e. confidence) highlights the need for calibration. 

When trained on Dataset II with clean and blurry images, as shown in Fig. \ref{fig:dataset2_uncalib}, there is a slight improvement in model performance, however, the uncalibrated model still has a tendency to over-report confidence. For this dataset, the model accuracy is 62\%, while the average confidence is 82\%. A similar trend is seen when training on Dataset III with clean and noisy images. In this case, the model accuracy is 68\%, with the average confidence being 78\%. Although the gap between model confidence and accuracy decreases with the calibration, we note that temperature scaling does not seem to perform as well on Dataset III relative to Datasets I and II. As can be observed from Fig. \ref{fig:dataset3_calib}, there is still a gap of 4\% between average confidence and accuracy in the calibrated model, whereas the gap is reduced to 1\% and 2\% for the Dataset I and II through calibration, respectively. 

%Of note, we observe that the model trained on Dataset IV (with clean, noisy, and blurry images) is already reasonably well-calibrated without post-processing, and temperature scaling is able to further bridge that 1\% gap as well, such that the reported model confidence is in complete agreement with its true accuracy. 

The reliability diagrams show that none of the models manage to achieve perfect calibration even after temperature scaling despite the average accuracy and confidence lining up. The achieved accuracies of the bins for the calibrated models still at least somewhat deviate from the ideal diagonal in all cases, although they are, on average, closer to this diagonal than their uncalibrated counterparts, therefore further supporting the advantage of employing a calibrated neural network.

In addition, to evaluate the model's capabilities to generalize at higher blur and noise levels, we plot the accuracy and confidence of the calibrated and uncalibrated model trained on each Dataset from I to III versus increasing blur and noise. We use a value of $\sigma$ ranging from 1 to 256 for blur and from 1 to 50 for noise, with a logarithmic step size for blur and a linear step for noise. 

\begin{figure}[t!]
    \centering
 \hspace*{-4mm}
 \includegraphics[scale=0.18]{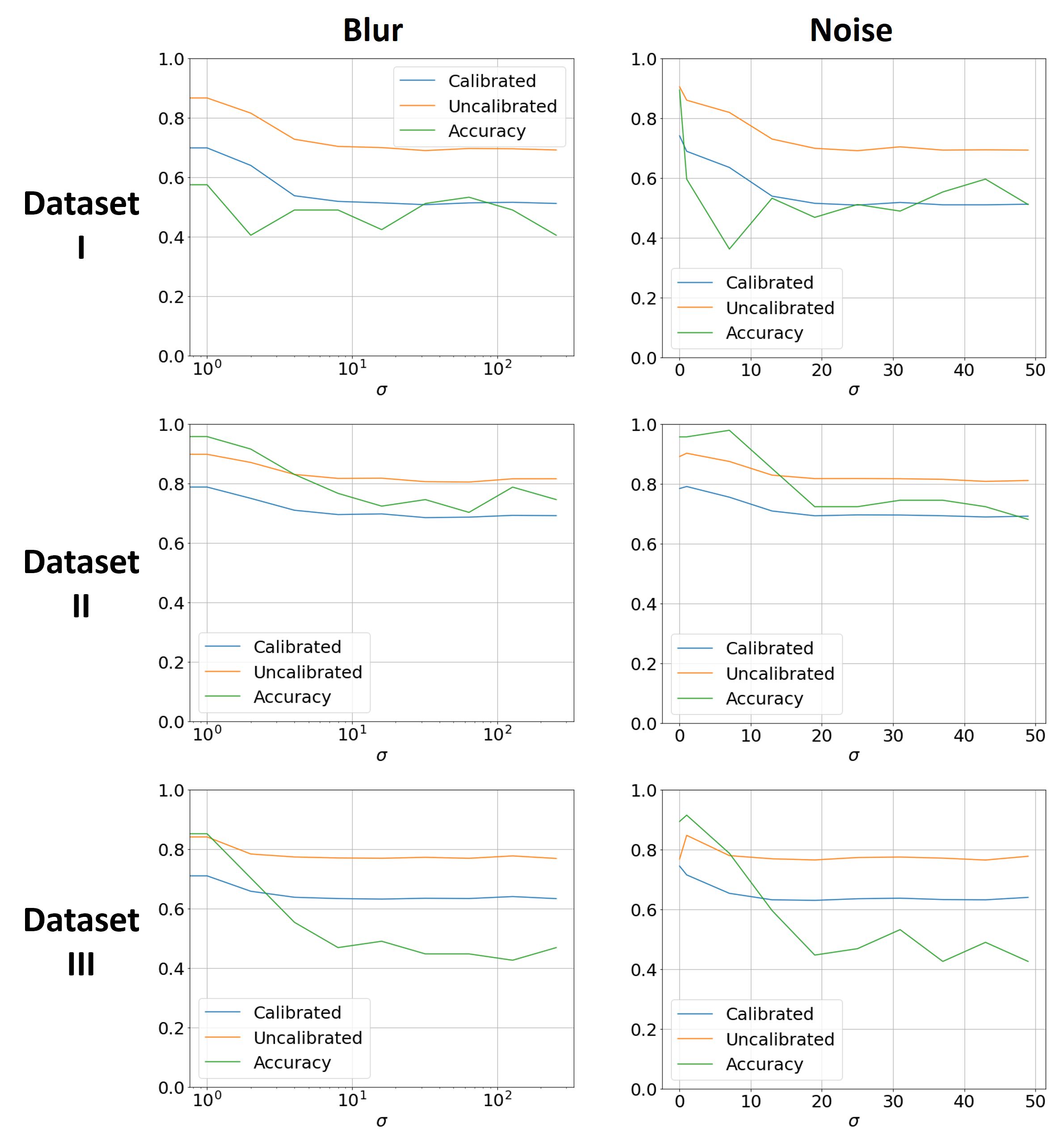}
    \caption{Results for model performance when subjected to increasing levels of blur and noise.}
    \label{fig:my_label}
    \vspace*{-4mm}
\end{figure}

The results of model performance versus increasing blur and noise levels are presented in Fig. \ref{fig:my_label}.
% \ref{fig:test1} to \ref{fig:test3}.
The model trained on Dataset I shows a sharp decrease in accuracy when noise and blur are introduced, even though the reported confidences remain relatively high. It is only at large levels of noise and blur that the average confidence and accuracy line up---at around the 50\% mark---but that is still too low to be of clinical significance. The accuracy drop for the model trained on Dataset II is much smaller, although there is still a significant gap between average confidence and accuracy for a significant portion of our testing range. The accuracy and confidence values converge at a higher point (75\%) than in the previous case. The model trained on Dataset III appears to be the worst performing of the three cases that have been considered since the accuracy drop occurs significantly for a low $\sigma$ and it never converges with the confidence. At the maximum blur and noise levels, which are beyond what may be encountered in a real-world setting, there is a significant gap between accuracy and confidence, which renders the model uninterpretable. 
%Finally, the model trained on Dataset IV remains calibrated and interpretable for a larger part of the testing range for noise and blur. The drop-off in accuracy only occurs at higher levels and has the least magnitude of the four trained models. We note that the model is more interpretable for blur than for noise. While there is no gap between confidence and accuracy as we increase blur until the drop-off, there is still a 10\% gap when testing with noise.   

As discussed previously, accuracy, precision, sensitivity, etc are insufficient metrics to determine the performance of a model. Keeping that in mind, we define the best-performing model in our tests to be the one that minimizes ACE, MCE, ECE, and the accuracy-confidence gap, while maximizing the aforementioned metrics. Additionally, the model should be able to remain calibrated even when exposed to noisy and/or blurry data.
%These criteria are fulfilled by the calibrated model trained on Dataset IV.
The model trained on Dataset I has the highest ACE, MCE, and ECE, which suggests its poor performance with regard to reliability. The model trained on Dataset III (noisy data) has the highest accuracy, however, it is miscalibrated even after temperature scaling. Additionally, it is unable to generalize over higher levels of noise and blur as discussed previously. The model trained on dataset II has lower accuracy, however, after calibration, the confidence estimates are close to ideal. The scalar metrics ACE, MCE, and ECE are also the lowest amongst the three models for this model, which makes it comparatively well-performing. Thus, considering these extra metrics allows us to choose a model that is not only accurate, but also reliable and interpretable.
In a clinical context, the best-performing model (i.e., the model trained on Dataset II) would produce a confidence for the predicted polyp class that is a representation of its true accuracy. A clinician would interpret the model's confidence as a reliable measure for encountering a particular class of CRC polyps---a prediction confidence of 80\% would be equivalent to an 80\% accuracy of the model--- which can be used to more reliably distinguish neoplastic polyp classes from their non-neoplastic counterparts. Thus, as opposed to the confidence reported by an uncalibrated network, a temperature scaled network incorporates a clinically-relevant significance to the model's classification confidence for a given HySenSe output.

\vspace*{-3mm}\section{CONCLUSIONS}
In this paper, we address the reliability and interpretability of our previously developed best-performing neural network  model in \cite{Venkatayogi2022PitPatternCO} by using a post-processing temperature scaling method for confidence calibration. Through testing via non-ideal inputs with blur and noise, we highlighted the difference in the confidence-accuracy gap of the model predictions for uncalibrated and calibrated models using reliability diagrams. We demonstrated that utilizing traditional metrics such as accuracy are not sufficient to encapsulate model performance demanding additional metrics to capture model reliability and interpretability for handling real-world scenarios. %We find that the best performing model in our tests to be the calibrated model trained on Dataset IV, which includes blurry and noisy images. 
Using the additional metrics, we show that the proposed confidence-calibration method can provide a better AI algorithm  for a reliable CRC polyp diagnosis and classification. Such AI algorithms can provide a trustworthy and reliable outputs to potentially  reduce the EDMR by making it easier for the clinician to take decision control for low confidence predictions. Our future works will primarily focus on the variance in this confidence estimate, which is encapsulated by another metric called uncertainty. 

\addtolength{\textheight}{-6cm}   % This command serves to balance the column lengths

%%%%%%%%%%%%%%%%%%%%%%%%%%%%%%%%%%%%%%%%%%%%%%%%%%%%%%%%%%%%%%%%%%%%%%%%%%%%%%%%

%%%%%%%%%%%%%%%%%%%%%%%%%%%%%%%%%%%%%%%%%%%%%%%%%%%%%%%%%%%%%%%%%%%%%%%%%%%%%%%%

%%%%%%%%%%%%%%%%%%%%%%%%%%%%%%%%%%%%%%%%%%%%%%%%%%%%%%%%%%%%%%%%%%%%%%%%%%%%%%%%

\bibliographystyle{IEEEtran}
\bibliography{ISMR2023_ML}

\end{document}